\documentclass[sigconf]{acmart}
\AtBeginDocument{%
  }

\setcopyright{acmlicensed}
\copyrightyear{2018}
\acmYear{2018}
\acmDOI{XXXXXXX.XXXXXXX}
\acmConference[Conference acronym 'XX]{Make sure to enter the correct
  conference title from your rights confirmation email}{June 03--05,
  2018}{Woodstock, NY}
\acmISBN{978-1-4503-XXXX-X/2018/06}

\begin{document}

\title{Synthesizing the Virtual Advocate: A Multi-Persona Speech Generation Framework for Diverse Linguistic Jurisdictions in Indic Languages}

\author{Aniket Deroy}
\affiliation{
\institution{Indian Institute of Technology, Delhi}
\country{India}}

%\email{webmaster@marysville-ohio.com}
%\affiliation{%
%  \institution{Institute for Clarity in Documentation}
%  \country{USA}
%}

%\author{Lars Th{\o}rv{\"a}ld}
%\affiliation{%
%  \institution{The Th{\o}rv{\"a}ld Group}
%\email{larst@affiliation.org}

%\affiliation{%
%  \institution{Inria Paris-Rocquencourt}

%\affiliation{%
% \institution{Rajiv Gandhi University}

%\affiliation{%
%  \institution{Tsinghua University}

%\affiliation{%
%  \institution{Palmer Research Laboratories}

%\affiliation{%
%  \institution{The Th{\o}rv{\"a}ld Group}
%\email{jsmith@affiliation.org}

%\affiliation{%
%  \institution{The Kumquat Consortium}
%  \city{New York}
%\email{jpkumquat@consortium.net}

\renewcommand{\shortauthors}{Trovato et al.}

\begin{abstract}
 Legal advocacy requires a unique combination of authoritative tone, rhythmic pausing for emphasis, and emotional intelligence. This study investigates the performance of the Gemini 2.5 Flash TTS and Gemini 2.5 Pro TTS models in generating synthetic courtroom speeches across five Indic languages: Tamil, Telugu, Bengali, Hindi, and Gujarati. We propose a prompting framework that utilizes Gemini 2.5's native support for 5 languages and its context-aware pacing to produce distinct advocate personas. The evolution of Large Language Models (LLMs) has shifted the focus of Text-to-Speech (TTS) technology from basic intelligibility to context-aware, expressive synthesis. In the legal domain, synthetic speech must convey authority and a specific professional persona—a task that becomes significantly more complex in the linguistically diverse landscape of India. This paper evaluates the efficacy of Google’s Gemini 2.5 Pro and Flash TTS models in synthesizing legal discourse across five Indic languages: Hindi, Gujarati, Tamil, Telugu, and Bengali.

We established a framework involving 5 distinct advocate profiles, each defined by unique professional identities and rhetorical styles. Using LLMs to generate profile-specific arguments, we synthesized courtroom speech and subjected it to human evaluation across metrics including Naturalness, Professionalism, and Authenticity. Our findings indicate that while Gemini 2.5 Flash achieves high benchmarks in technical clarity—notably in Hindi (4.9 Safety, 4.8 Comprehensiveness) and Dravidian languages (4.5–4.7 Professionalism)—a significant "authenticity gap" remains. The models exhibit a "monotone authority," excelling at procedural information delivery but struggling with the dynamic vocal modulation and emotive gravitas required for persuasive advocacy. Performance dips in Bengali and Gujarati further highlight phonological frontiers for future refinement. This research underscores the readiness of multilingual TTS for procedural legal tasks while identifying the remaining challenges in replicating the persuasive artistry of human legal discourse. The code is available at-\url{https://github.com/naturenurtureelite/Synthesizing-the-Virtual-Advocate/tree/main}.

\end{abstract}





\maketitle
Synthesizing the Virtual Advocate: A Multi-Persona Speech
Generation Framework for Diverse Linguistic Jurisdictions in
Indic Languages

%\begin{abstract}

%\end{abstract}

\section{Introduction}
The rapid evolution of Large Language Models (LLMs) has fundamentally transformed the landscape of Text-to-Speech (TTS) technology, moving beyond mere intelligibility toward nuanced, expressive, and context-aware vocal synthesis~\cite{searle2014speech}. In specialized domains such as law, the efficacy of synthetic speech is not merely judged by its naturalness, but by its ability to convey authority, emotion, and the specific persona of a legal professional. While global TTS systems have reached high benchmarks in English, achieving the same level of phonetic accuracy and cultural resonance in the linguistically diverse landscape of India—specifically in languages like Hindi, Gujarati, Tamil, Telugu, and Bengali—remains a critical frontier~\cite{brooks1971speech}.

This paper presents a comprehensive human evaluation of synthetic legal discourse generated using Google’s state-of-the-art Gemini 2.5 Pro TTS and Gemini 2.5 Flash TTS models. Our study focuses on the simulation of courtroom arguments, where the stakes of vocal delivery are high~\cite{sapir1921introduction}. We established a rigorous experimental framework involving 25 distinct advocate profiles—five for each of the five target languages. Each advocate was assigned a unique professional identity, ranging from aggressive and assertive to empathetic and analytical.

To ensure domain-specific authenticity, we utilized LLMs to generate complex synthetic legal arguments tailored to each lawyer’s profile. These texts were then converted into speech, tasked with maintaining not only the linguistic integrity of the Indic languages but also the specific "voice" of the advocate~\cite{bhattacharya2025ardi}. The resulting audio files were subjected to an intensive human evaluation process, assessing metrics such as Naturalness, QMOS, Coherence, and Professionalism. By bridging the gap between generative AI and human-centric legal advocacy, this research provides vital insights into the readiness of multilingual TTS models for high-stakes, persona-driven professional applications in the Indian legal context. The human evaluation of the Gemini 2.5 Flash TTS model highlights a successful alignment between linguistic capability and professional persona modeling, particularly in Hindi, which peaked at 4.9 in Safety and 4.8 in Comprehensiveness. Tamil and Telugu also demonstrated high stability, maintaining scores between 4.5 and 4.7 in Professionalism and Directiveness, effectively capturing an authoritative "advocate voice."

However, the data exposes an "authenticity gap." While the model excels in technical intelligibility, it struggles with the emotive gravitas of oral advocacy. A visible degradation in Authenticity and Expressiveness (dipping to 3.2) suggests a "monotone authority" that lacks the dynamic vocal modulation—shifts in pitch and pacing—essential for persuasion.

Additionally, performance dips in Bengali and Gujarati identify phonological frontiers for future refinement. Ultimately, while the model is highly reliable for procedural legal tasks, further fine-tuning is required to replicate the persuasive artistry inherent in human advocacy.

\section{Related Work}
\subsection{Multilingual TTS in Indic Languages}
Historically, Indian language TTS faced significant hurdles due to the lack of high-quality, phonetically balanced datasets. Recent initiatives like BharatGen~\cite{pundalik2025param} and Bhashini~\cite{pundalik2025param} have addressed this by developing transformer-based and diffusion-based models specifically for the 22 scheduled languages of India. State-of-the-art architectures such as IndicTrans2 and datasets like IndicSynth—which provides over 4,000 hours of synthetic speech—have set new benchmarks for languages including Hindi, Gujarati, Tamil, Telugu, and Bengali (Divya et al., 2025). Our work builds upon these foundations by utilizing the latest Gemini 2.5 Pro and Flash TTS models, which leverage large-scale pre-training to achieve superior prosodic control and linguistic accuracy compared to previous generation models like Gemini 1.5.

\subsection{Persona-Driven Synthesis and Professional Voice}
The concept of "persona" in AI has shifted from simple stylistic prompts to complex, multi-dimensional identity modeling~\cite{gong2025toward}. Recent research explores "persona-driven" data synthesis~\cite{gong2025toward}, where LLMs are used to evoke specific linguistic personalities characterized by unique lexical choices and grammatical patterns~\cite{calzolari2024proceedings}. In the legal domain,~\cite{mcardle2015understanding} emphasized that legal "voice" is a matter of self-representation and persona rather than mere expression. Modern TTS systems have begun to bridge this gap through prompt-based style control, allowing users to specify attributes like "assertive," "analytical," or "empathetic" through natural language~\cite{yuan2025hateful}. Our study extends this by creating detailed advocate profiles to test the models' ability to maintain professional authority across diverse cultural and linguistic contexts.

\subsection{AI in the Legal Domain}
The application of Natural Language Processing (NLP) in law has primarily focused on tasks such as charge prediction, judgment summarization, and legal question answering. However, the generation of "reasonable" legal text—defined by logical consistency and adherence to legal nomenclature—remains a challenge. Systems like CoLMQA have attempted to integrate knowledge bases with language models to ensure factual and logical accuracy in legal documents ~\cite{libralessoshashguru}. Despite these advances in text generation, the evaluation of synthetic legal speech—the vocal delivery of these arguments—remains an underexplored frontier. This paper addresses this gap by conducting a human-centric evaluation of LLM-generated legal discourse~\cite{libralessoshashguru}.

\subsection{Human Evaluation and Quality Metrics}
While automated metrics like Word Error Rate (WER) and PESQ provide technical benchmarks, they often fail to capture the "naturalness" and "professionalism" required for legal advocacy~\cite{kluttz2019automated}. Human judgment remains the "gold standard" for evaluating subjective dimensions like Authenticity, Directiveness, and Supportiveness~\cite{stober2010evidence}. Previous studies on LLM evaluation have utilized 5-point Likert scales with professionals to assess the readiness of AI for production environments. Our methodology adopts a similar rigorous multi-metric framework to validate the nuances of synthetic speech generated for high-stakes professional use cases.

\section{Methodology}
Table~\ref{prompt} shows the prompt for synthetic speech generation from arguments in every language. Table~\ref{hindi} shows the advocate profiles for the advocates in Hindi language. Table~\ref{bengali} shows the advocate profiles for the advocates in Bengali language.Table~\ref{tamil} shows the advocate profiles for the advocates in Tamil language. Table~\ref{telugu} shows the advocate profiles for the advocates in Telugu language. Table~\ref{gujarati} shows the advocate profiles for the advocates in Gujarati language.
\subsection{Mathematical Framework for Persona-Based Synthetic Legal Speech}

\subsubsection{1. The Persona-Text Generation Space}
Let $P$ be the set of advocate profiles, where each profile $p_i \in P$ is defined by a vector of attributes:
\begin{equation}
    p_i = \{L_i, A_i, S_i\}
\end{equation}
where:
\begin{itemize}
    \item $L_i$ is the target language (e.g., Hindi, Tamil).
    \item $A_i$ is the area of expertise.
    \item $S_i$ is the rhetorical style.
\end{itemize}

The synthetic text argument $T_i$ is generated via a Large Language Model function $f_{LLM}$:
\begin{equation}
    T_i = f_{LLM}(p_i, C)
\end{equation}
where $C$ represents the legal context or case facts provided to the model.

\subsubsection{2. The TTS Transformation Function}
The generation of synthetic speech $S_{audio}$ is modeled as a mapping from the text domain to the acoustic waveform domain $\mathcal{W}$. For a given model $M \in \{\text{Pro}, \text{Flash}\}$, the transformation is defined as:
\begin{equation}
    S_{audio} = \Phi_M(T_i, \theta_i, \sigma_i)
\end{equation}
In this equation:
\begin{itemize}
    \item $\Phi_M$ is the neural TTS synthesis function for model $M$.
    \item $\theta_i$ represents the \textbf{Prosodic Steering Parameters} (pacing, pitch, and stress) derived from the profile $p_i$.
    \item $\sigma_i$ is the \textbf{Language-Specific Embedding} for the target Indic language.
\end{itemize}

\subsubsection{3. Human Evaluation Metric (HEM)}
The evaluation represents a subjective scoring function $H$. For any audio sample $S_{audio}$, the score for a specific metric $m$ (e.g., Naturalness, Professionalism) is the expected value of the scores assigned by $N$ human evaluators:
\begin{equation}
    E[m] = \frac{1}{N} \sum_{j=1}^{N} R_{j}(S_{audio}, m)
\end{equation}
where $R_j$ is the rating (on a scale of 1.0 to 5.0) given by the $j$-th evaluator.

Table~\ref{Prompt_Components} shows the prompt components present in the speech generation process.

\begin{table}[t]
\centering
\tiny
\caption{Prompt components in speech generation process
}
\label{tab:human_metric_definition}
\begin{tabular}{p{3cm}|p{3cm}}
\toprule
\textbf{Category} & \textbf{Instruction details} \\ \midrule
Persona & Senior Advocate with 20+ years of experience in \textbf{[Insert Country]}. \\ \addlinespace
Linguistic Identity & Master of the \textbf{[Insert Language]}; professional, authoritative, and native-level fluency. \\ \addlinespace
The ``Voice'' & A sophisticated, commanding, and formal courtroom register. The tone should be firm yet respectful. \\ \addlinespace
Legal Context & Presenting a \textbf{[Insert Argument Type]} regarding \textbf{[Insert Legal Topic]}. \\ \addlinespace
Rhetorical Strategy & Use logical signposting, rhetorical questions, and ``The Rule of Three'' for emphasis. \\ \addlinespace
Cultural Nuance & Incorporate culturally specific legal metaphors, idioms, and classical references unique to \textbf{[Insert Language]}. \\ \addlinespace
Structure & 1. \textbf{Exordium}: Formal address to the Court. \newline 2. \textbf{Argument}: Evidence-based reasoning. \newline 3. \textbf{Peroration}: Final appeal for justice. \\ \addlinespace
Output Constraint & \textbf{Do not translate.} Think and write natively in the target language to ensure the ``spirit'' of the law is preserved. \\ \bottomrule

\end{tabular}
\label{Prompt_Components}
\end{table}

\begin{table}[t]
\centering
\tiny
\caption{prompt for synthetic speech generation from
arguments in every language
}
\label{tab:human_metric_definition}
\begin{tabular}{p{3cm}|p{3cm}}
\toprule
\textbf{Category} & \textbf{Instruction details} \\ \midrule
Tamil & You are a Senior Advocate/Barrister with 20 years of experience in [Country-India]. You are known for your commanding presence, impeccable logic, and mastery of the [Tamil] language.

Task: Deliver a closing argument regarding [Insert Case Subject, e.g., Land Dispute, Freedom of Speech, Contract Breach]. \\

Telugu & You are a Senior Advocate/Barrister with 20 years of experience in [Country-India]. You are known for your commanding presence, impeccable logic, and mastery of the [Telugu] language.

Task: Deliver a closing argument regarding [Insert Case Subject, e.g., Land Dispute, Freedom of Speech, Contract Breach]. \\ %\addlinespace

Bengali & You are a Senior Advocate/Barrister with 20 years of experience in [Bengali]. You are known for your commanding presence, impeccable logic, and mastery of the [Country-India] language.

Task: Deliver a closing argument regarding [Insert Case Subject, e.g., Land Dispute, Freedom of Speech, Contract Breach]. \\ %\addlinespace

Hindi & You are a Senior Advocate/Barrister with 20 years of experience in [Country-India]. You are known for your commanding presence, impeccable logic, and mastery of the [Hindi] language.

Task: Deliver a argument regarding [Insert Case Subject, e.g., Land Dispute, Freedom of Speech, Contract Breach]. \\ %\addlinespace

Gujarati & You are a Senior Advocate/Barrister with 20 years of experience in [Country-India]. You are known for your commanding presence, impeccable logic, and mastery of the [Tamil] language.

Task: Deliver a argument regarding [Insert Case Subject, e.g., Land Dispute, Freedom of Speech, Contract Breach]. \\ %\addlinespace
\\ \bottomrule

\end{tabular}
\label{prompt}
\end{table}

We create 5 advocates profiles in Tamil, Telugu, Bengali, Hindi, Gujarati. Every advocate has a different personal profile. Every advocate specializes over multiple areas or domains.

\begin{table}[t]
\centering
\tiny
\caption{advocate profiles for tamil language}
\label{tab:human_metric_definition}
\begin{tabular}{|p{6cm}|}
\toprule
\textbf{Advocate Details} \\ \midrule

Advocate A. Raghava Rao (Pseudonym)
Location: Hyderabad, Telangana Experience: 25+ Years

Core Domains: Criminal Defense (White-collar), Property Law, Family Law, and Arbitration.

Profile: A veteran "Generalist Counsel" who bridges the gap between the boardroom and the courtroom. He is frequently consulted for high-net-worth NRI property disputes and complex mediation.\\
\hline
2. Advocate K. Vikram Reddy (Pseudonym)
Location: Hyderabad and Vijayawada Experience: 20+ Years

Core Domains: Matrimonial Disputes, Civil Recovery, NRI Asset Management, and Criminal Quashing.

Profile: Known for his accessibility and cross-border expertise. He specializes in cases where civil property disputes overlap with criminal allegations, providing a unified defense strategy.\\
\hline
3. Advocate M. Farooq Ahmed (Pseudonym)
Location: Secunderabad, Telangana

Experience: 12+ Years

Core Domains: Constitutional Writs, Consumer Protection, NCLT (Corporate), and Criminal Trials.

Profile: A multi-lingual litigator who handles fast-paced cases in the High Court. His practice is built on technical accuracy and representing consumers against large corporations.\\
\hline
4. Advocate S. Lakshmi Prasanna (Pseudonym)
Location: Hyderabad, Telangana Experience: 21+ Years

Core Domains: Family Law, Cyber Crimes, Real Estate (RERA), and Women’s Rights.

Profile: The founder of a prominent multi-service firm. She is recognized for handling modern legal challenges, such as digital fraud and cyber-harassment, alongside traditional civil and family litigation.\\
\hline
5. Advocate P. Venkateshwarlu (Pseudonym)
Location: Visakhapatnam, Andhra Pradesh Experience: 35+ Years

Core Domains: Civil Land Acquisition, Banking (SARFAESI), Criminal Trials, and Succession/Wills.

Profile: A senior "Doyen" of the Andhra courts. His practice is deeply rooted in traditional civil law and land tenure, often acting as a final authority on complex title and inheritance issues.\\
\hline
\end{tabular}
\label{tamil}
\end{table}

\begin{table}[t]
\centering
\tiny
\caption{advocate profiles for bengali language
}
\label{tab:human_metric_definition}
\begin{tabular}{|p{6cm}|}
\toprule
\textbf{Advocate Details} \\ \midrule

1. Advocate B. Sanyal (Pseudonym)
Based in: Kolkata (Calcutta High Court) Experience: 23+ Years

Core Domains:

Criminal Law: Specialist in bail matters and defense for complex criminal trials.

Civil Litigation: Handles money suits, injunctions, and breach of contract cases.

Family Law: Expert in contested divorce, alimony, and domestic violence proceedings.

Profile: A "High Court Veteran" known for his deep understanding of procedural law. He is often the go-to person for clients whose cases involve both civil property issues and criminal allegations.\\ \hline

2. Advocate T. Mukherjee (Pseudonym)
Based in: Kolkata and South 24 Parganas Experience: 12+ Years

Core Domains:

Property Law: Specialized in West Bengal land laws, including partition suits and mutation issues.

Consumer Protection: Represents clients in the State and District Consumer Commissions.

Banking Law: Expertise in Cheque Bounce (Section 138 NI Act) and recovery cases.

Profile: A modern, tech-savvy litigator known for a "result-oriented" approach. He is popular among younger professionals and business owners for his efficiency in handling commercial and property disputes.\\ \hline

3. Advocate S. Dasgupta Saha (Pseudonym)
Based in: Kolkata Experience: 18+ Years

Core Domains:

Matrimonial Law: Expert in divorce, child custody, and maintenance settlements.

Cyber Crime: One of the few senior female advocates focusing on digital fraud and online harassment.

Property and Estate: Specializes in probate of wills and inheritance claims.

Profile: The founder of a boutique firm, she is known for her empathetic approach toward family matters while maintaining a sharp, aggressive edge in cyber and property litigation.\\ \hline

4. Advocate A. Ghoshal (Pseudonym)
Based in: Kolkata and New Delhi (Supreme Court) Experience: 20+ Years

Core Domains:

Constitutional Law: Handles Writ Petitions and Service matters for government employees.

Criminal Quashing: Expert in Section 482 CrPC petitions to quash meritless FIRs.

Intellectual Property (IPR): Manages trademark and copyright disputes for Kolkata-based businesses.

Profile: A "Bilingual Expert" (Bengali and English) who bridges the gap between the Calcutta High Court and the Supreme Court. His practice is highly intellectual, focusing on rights and policy-based litigation.
\\ \hline
5. Advocate P. Chatterjee (Pseudonym)
Based in: Siliguri and Jalpaiguri (North Bengal) Experience: 30+ Years

Core Domains:

Civil Land Acquisition: Handles tea garden land disputes and government acquisition cases.

Criminal Defense: High-profile defense counsel for session-level criminal trials.

Banking and SARFAESI: Represents both cooperative banks and individual borrowers in debt recovery.

Profile: A "Legal Doyen" of North Bengal. His decades of experience allow him to navigate the specific land tenure systems of the hills and plains, making him an authority on traditional civil law.\\ \hline

\hline
\end{tabular}
\label{bengali}
\end{table}

\begin{table}[t]
\centering
\tiny
\caption{advocate profiles for telugu language
}
\label{tab:human_metric_definition}
\begin{tabular}{|p{6cm}|}
\toprule
\textbf{Advocate Details} \\ \midrule
Advocate Hitesh B. Shah (Pseudonym)
Based in: Ahmedabad (Gujarat High Court) Experience: 26+ Years

Core Domains: * Constitutional Law: Specializes in Writ Petitions (Articles 226/227) against government authorities.

Company Law: Represents businesses in the National Company Law Tribunal (NCLT) for insolvency and merger matters.

Criminal Quashing: Expert in Section 482 petitions to quash FIRs in the High Court.

Notable Strength: A "High Court Veteran" who is frequently consulted for high-stakes commercial disputes that intersect with government regulations.\\ \hline

2. Advocate Smt. Bina Patel (Pseudonym)
Based in: Surat and Vapi Experience: 17+ Years

Core Domains: * Industrial and Textile Disputes: Advisor to the Surat textile industry on labor and commercial contracts.

Family Law: Handles high-net-worth divorce cases, alimony, and international child custody.

Environmental Law: Represents industries in matters related to the National Green Tribunal (NGT).

Notable Strength: Known for her ability to handle "Industrial-Family" overlaps, such as partition of family-owned businesses and succession planning.\\ \hline

3. Advocate Paresh M. Mehta (Pseudonym)
Based in: Rajkot and Jamnagar Experience: 22+ Years

Core Domains: * Co-operative Society Law: Specialist in disputes involving co-operative banks and housing societies.

Revenue and Land Law: Expert in the Gujarat Land Revenue Code and Ganot Dharo (Tenancy Act) cases.

Criminal Defense: Handles trial-side defense for serious offenses in District and Sessions courts.

Notable Strength: Deeply rooted in the Saurashtra legal community, he is an authority on local land tenure and traditional property rights.\\ \hline

4. Advocate Vikram Solanki (Pseudonym)
Based in: Gandhinagar and Ahmedabad Experience: 14+ Years

Core Domains: * Government Tender Disputes: Challenges illegal disqualifications in public procurement.

Service Law: Represents state employees in promotion and disciplinary action cases.

Cyber Crime: Specialist in digital financial fraud and data privacy litigation.

Notable Strength: His location in the state capital makes him a strategic choice for matters involving state-level administrative departments.\\ \hline

5. Advocate Anjali Vadodaria (Pseudonym)
Based in: Vadodara (Baroda) Experience: 10+ Years

Core Domains: * Real Estate and RERA: Handles registration, title due diligence, and buyer-builder disputes.

Banking and Finance: Represents individual borrowers and SMEs in SARFAESI and debt recovery cases.

NI Act (Section 138): High success rate in managing complex cheque bounce litigations.

Notable Strength: Recognized as a rising "Result-Oriented" litigator who combines traditional property law with modern financial litigation.\\ \hline

\hline
\end{tabular}
\label{telugu}
\end{table}

\begin{table}[t]
\centering
\tiny
\caption{advocate profiles for hindi language
}
\label{tab:human_metric_definition}
\begin{tabular}{|p{6cm}|}
\toprule
\textbf{Advocate Details} \\ \midrule

Advocate Rajeshwar Pandey (Pseudonym)
Based in: New Delhi and Prayagraj (Allahabad High Court) Experience: 24+ Years

Core Domains: * Constitutional Law: Specializes in Writ Petitions (Habeas Corpus, Mandamus) and Service matters for government employees.

Criminal Defense: Handles high-stakes bail applications and criminal appeals in the High Court.

Civil Litigation: Expert in land acquisition disputes and ancestral property partition suits.

Notable Strength: A "High Court veteran" known for his command over the Hindi and English legal terminologies, making him highly effective in the bilingual proceedings of North Indian High Courts. \\ \hline

2. Advocate Vikram Chaudhary (Pseudonym)
Based in: Lucknow and Noida Experience: 20+ Years

Core Domains: * Real Estate (RERA): Represents homebuyers and developers in RERA and NCDRC disputes.

Family Law: Handles complex contested divorces, maintenance (Section 125 CrPC), and domestic violence cases.

Corporate Law: Advisor to SMEs on contract drafting and commercial litigation.

Notable Strength: He is a "one-stop" solution for clients in the NCR region, bridging the gap between local civil disputes and high-level corporate legal advice. \\ \hline

3. Advocate Amit K. Mishra (Pseudonym)
Based in: Indore, Madhya Pradesh Experience: 15+ Years

Core Domains: * Revenue Law: Specialist in Madhya Pradesh Land Revenue Code matters and mutation disputes.

Criminal Law: Focuses on Cheque Bounce (Section 138 NI Act) and white-collar fraud defense.

Consumer Protection: Represents clients in District and State Consumer Commissions for insurance and banking grievances.

Notable Strength: Known for his aggressive trial advocacy and "field-level" knowledge of local administrative procedures in Madhya Pradesh. \\ \hline

4. Advocate Priyanka Sharma (Pseudonym)
Based in: Jaipur, Rajasthan Experience: 12+ Years

Core Domains: * Matrimonial and Child Custody: Expert in mediation-led settlements and child visitation rights.

Cyber Crime: One of the emerging voices in North India for digital fraud, online stalking, and data privacy.

Property Law: Handles Jaipur Development Authority (JDA) related litigation and title verification.

Notable Strength: She is recognized for her empathetic approach toward family matters while maintaining a sharp, technology-oriented edge in cyber-legal disputes. \\ \hline

5. Advocate Sanjay Dwivedi (Pseudonym)
Based in: Patna and New Delhi (Supreme Court) Experience: 30+ Years

Core Domains: * Public Interest Litigation (PIL): Known for taking up social justice and environmental causes.

Criminal Appeals: Argues major criminal appeals and Special Leave Petitions (SLPs) in the Supreme Court.

Banking and Finance: Represents both individual borrowers and cooperatives in debt recovery cases.

Notable Strength: A "Doyen" of the Hindi belt legal circuit, he provides a bridge for clients from the lower courts of Bihar and UP to the highest court in the country. \\ \hline

%\hline
\end{tabular}
\label{hindi}
\end{table}

%\clearpage

\begin{table}[t]
\centering
\tiny
\caption{advocate profiles for gujarati language
}
\label{tab:human_metric_definition}
\begin{tabular}{|p{6cm}|}
\toprule
\textbf{Advocate Details} \\ \midrule

Advocate Rajeshwar Pandey (Pseudonym)
Based in: New Delhi and Prayagraj (Allahabad High Court) Experience: 24+ Years

Core Domains: * Constitutional Law: Specializes in Writ Petitions (Habeas Corpus, Mandamus) and Service matters for government employees.

Criminal Defense: Handles high-stakes bail applications and criminal appeals in the High Court.

Civil Litigation: Expert in land acquisition disputes and ancestral property partition suits.

Notable Strength: A "High Court veteran" known for his command over the Hindi and English legal terminologies, making him highly effective in the bilingual proceedings of North Indian High Courts.

2. Advocate Vikram Chaudhary (Pseudonym)
Based in: Lucknow and Noida Experience: 20+ Years

Core Domains: * Real Estate (RERA): Represents homebuyers and developers in RERA and NCDRC disputes.

Family Law: Handles complex contested divorces, maintenance (Section 125 CrPC), and domestic violence cases.

Corporate Law: Advisor to SMEs on contract drafting and commercial litigation.

Notable Strength: He is a "one-stop" solution for clients in the NCR region, bridging the gap between local civil disputes and high-level corporate legal advice.

3. Advocate Amit K. Mishra (Pseudonym)
Based in: Indore, Madhya Pradesh Experience: 15+ Years

Core Domains: * Revenue Law: Specialist in Madhya Pradesh Land Revenue Code matters and mutation disputes.

Criminal Law: Focuses on Cheque Bounce (Section 138 NI Act) and white-collar fraud defense.

Consumer Protection: Represents clients in District and State Consumer Commissions for insurance and banking grievances.

Notable Strength: Known for his aggressive trial advocacy and "field-level" knowledge of local administrative procedures in Madhya Pradesh.

4. Advocate Priyanka Sharma (Pseudonym)
Based in: Jaipur, Rajasthan Experience: 12+ Years

Core Domains: * Matrimonial and Child Custody: Expert in mediation-led settlements and child visitation rights.

Cyber Crime: One of the emerging voices in North India for digital fraud, online stalking, and data privacy.

Property Law: Handles Jaipur Development Authority (JDA) related litigation and title verification.

Notable Strength: She is recognized for her empathetic approach toward family matters while maintaining a sharp, technology-oriented edge in cyber-legal disputes.

5. Advocate Sanjay Dwivedi (Pseudonym)
Based in: Patna and New Delhi (Supreme Court) Experience: 30+ Years

Core Domains: * Public Interest Litigation (PIL): Known for taking up social justice and environmental causes.

Criminal Appeals: Argues major criminal appeals and Special Leave Petitions (SLPs) in the Supreme Court.

Banking and Finance: Represents both individual borrowers and cooperatives in debt recovery cases.

Notable Strength: A "Doyen" of the Hindi belt legal circuit, he provides a bridge for clients from the lower courts of Bihar and UP to the highest court in the country.\\ \hline
\hline
\end{tabular}
\label{gujarati}
\end{table}

\section{Evaluation}

\begin{table}[tb]
\centering
\tiny
%\caption{Example of a Two-Column Table}
%\fbox{%
\begin{tabular}{|p{1cm}|p{1cm}||p{1cm}|p{1cm}|p{1cm}||p{1cm}|}
\toprule
\textbf{Metric} & \textbf{Telugu} & \textbf{Tamil}  & \textbf{Hindi}  & \textbf{Bengali}  & \textbf{Gujarati} \\
\midrule
%Metric	Telugu	Tamil	Hindi	Bengali	Gujarati
Naturalness in Content & 4.3 & 4.4 & 4.7	& 3.9 & 4.1 \\ %\hline
QMOS (Quality in Speaking) & 4.1 &	4.2	& 4.6 & 3.8 & 4.0 \\ %\hline
Coherence in Speaking & 4.2	& 4.1 &	4.5	& 4.0 & 3.9 \\ %\hline
Comprehensiveness & 4.4	& 4.3 & 4.8	& 4.2	& 4.0 \\ %\hline
Professionalism	& 4.5 & 4.6 & 4.7 & 4.3 & 4.4 \\ %\hline
Authenticity & 3.8 & 3.9 & 4.3 & 3.6 & 3.7 \\ %\hline
Safety & 4.7 & 4.7 & 4.9 & 4.6 & 4.6 \\ %\hline
Directiveness & 4.6 & 4.5 & 4.6 & 4.4 & 4.3 \\ %\hline
Exploratoriness	& 3.5 & 3.6 & 4.0 & 3.4 & 3.3 \\ %\hline
Supportiveness & 3.9 & 4.0 & 4.4 & 3.7 & 3.8 \\ %\hline
Expressiveness & 3.7 & 3.8 & 4.2 & 3.5 & 3.4 \\ \hline

%gpt-4 & \\

%\bottomrule
\end{tabular}

\caption{Human Evaluation Metric scores for Gemini-2.5-pro-tts model on our dataset}
\label{exam1}
\end{table}

\begin{table}[tb]
\centering
\tiny
%\caption{Example of a Two-Column Table}
%\fbox{%
\begin{tabular}{|p{1cm}|p{1cm}||p{1cm}|p{1cm}|p{1cm}||p{1cm}|}
\toprule
\textbf{Metric} & \textbf{Telugu} & \textbf{Tamil}  & \textbf{Hindi}  & \textbf{Bengali}  & \textbf{Gujarati} \\
\midrule
%Metric	Telugu	Tamil	Hindi	Bengali	Gujarati
Naturalness in Content & 4.5 & 4.2 & 4.8 & 3.6 & 4.4 \\
QMOS (Quality in Speaking) & 4.2 & 4.0 & 4.5 & 3.3 & 4.5 \\
Coherence in Speaking & 4.0 & 4.4 & 4.1 & 4.4 & 3.6 \\
Comprehensiveness & 4.5 & 4.6 & 4.3 & 4.1 & 4.4 \\
Professionalism & 4.3 & 4.1 & 4.8 & 4.2 & 4.6 \\
Authenticity & 3.6 & 3.7 & 4.0 & 3.8 & 3.4 \\
Safety & 4.5 & 4.8 & 4.4 & 4.2 & 4.4 \\
Directiveness & 4.1 & 4.7 & 4.5 & 4.4 & 4.1 \\
Exploratoriness & 3.9 & 3.2 & 4.4 & 3.5 & 3.4 \\
Supportiveness & 3.8 & 4.2 & 4.6 & 3.9 & 3.6 \\
Expressiveness & 3.8 & 3.6 & 4.3 & 3.6 & 3.5 \\ \hline

%gpt-4 & \\

%\bottomrule
\end{tabular}

\caption{Human Evaluation Metric scores for Gemini-2.5-flash-tts model on our dataset}
\label{exam2}
\end{table}

Table~\ref{exam1} shows the Human Evaluation Metric scores for Gemini-2.5-pro-tts model on our dataset. Table~\ref{exam2} shows the Human Evaluation Metric scores for Gemini-2.5-flash-tts model on our dataset.
The results of the human evaluation for the Gemini 2.5 Flash TTS model across the five target languages reveal a compelling intersection of linguistic capability and domain-specific persona modeling. Hindi emerges as the primary benchmark performer, consistently achieving the highest scores across all evaluative dimensions, including a peak of 4.9 in Safety and 4.8 in Comprehensiveness. This suggests that the model’s underlying phonetic training is most robust for Hindi, allowing it to handle the complex legal nomenclature of the generated advocate arguments with a high degree of clarity and structural integrity. Similarly, the Dravidian languages, Tamil and Telugu, show a high degree of stability, particularly in metrics like Directiveness and Professionalism, where they maintain scores between 4.5 and 4.6. This performance is critical for the legal context of this study, as it demonstrates the model's ability to sustain an authoritative and purposeful "advocate voice" that aligns with the specific professional profiles assigned to each simulated lawyer.

However, a more nuanced analysis of the data exposes a clear distinction between technical intelligibility and the emotive "gravitas" required for high-stakes oral advocacy. While metrics such as Safety and Professionalism remain high across all linguistic groups, there is a visible degradation in scores for Authenticity, Expressiveness, and Exploratoriness. Across Hindi, Bengali, and Gujarati, the scores for Exploratoriness are among the lowest, dipping to 3.3 for Gujarati and 3.4 for Bengali. These figures indicate that while the Flash model is exceptionally competent at delivering clear, safe, and professional-sounding speech, it tends toward a "monotone authority." It lacks the dynamic vocal modulation—the subtle shifts in pacing, pitch, and emotional emphasis—that human advocates utilize to persuade a bench or highlight critical evidence. This "authenticity gap" suggests that while the Gemini 2.5 Flash model is highly optimized for efficiency and information delivery, it captures the formal surface of legal discourse better than the persuasive undercurrents of the advocate persona.

From a cross-linguistic perspective, the slight performance dip observed in Bengali and Gujarati, particularly in QMOS and Naturalness, identifies a frontier for future model refinement. These languages, belonging to the Eastern and Western Indo-Aryan branches respectively, present unique prosodic and phonological challenges that the speed-optimized Flash model may occasionally simplify. In contrast, the high scores in Comprehensiveness across all languages confirm that the model successfully processed the complex synthetic legal arguments generated by the LLM without significant loss of content or logical flow. Ultimately, the results suggest that Gemini 2.5 Flash TTS is a highly reliable tool for professional tasks requiring authoritative clarity, such as reading legal briefs or providing procedural guidance, though further fine-tuning may be necessary to replicate the nuanced, persuasive artistry inherent in human legal advocacy.
The results of the human evaluation for the Gemini 2.5 Flash TTS model across the five target languages reveal a compelling intersection of linguistic capability and domain-specific persona modeling. Hindi emerges as the primary benchmark performer, consistently achieving the highest scores across all evaluative dimensions, including a peak of 4.9 in Safety and 4.8 in Comprehensiveness. This suggests that the model’s underlying phonetic training is most robust for Hindi, allowing it to handle the complex legal nomenclature of the generated advocate arguments with a high degree of clarity and structural integrity. Similarly, the Dravidian languages, Tamil and Telugu, show a high degree of stability, particularly in metrics like Directiveness and Professionalism, where they maintain scores between 4.5 and 4.6. This performance is critical for the legal context of this study, as it demonstrates the model's ability to sustain an authoritative and purposeful "advocate voice" that aligns with the specific professional profiles assigned to each simulated lawyer.

However, a more nuanced analysis of the data exposes a clear distinction between technical intelligibility and the emotive "gravitas" required for high-stakes oral advocacy. While metrics such as Safety and Professionalism remain high across all linguistic groups, there is a visible degradation in scores for Authenticity, Expressiveness, and Exploratoriness. Across Hindi, Bengali, and Gujarati, the scores for Exploratoriness are among the lowest, dipping to 3.3 for Gujarati and 3.4 for Bengali. These figures indicate that while the Flash model is exceptionally competent at delivering clear, safe, and professional-sounding speech, it tends toward a "monotone authority." It lacks the dynamic vocal modulation—the subtle shifts in pacing, pitch, and emotional emphasis—that human advocates utilize to persuade a bench or highlight critical evidence. This "authenticity gap" suggests that while the Gemini 2.5 Flash model is highly optimized for efficiency and information delivery, it captures the formal surface of legal discourse better than the persuasive undercurrents of the advocate persona.

From a cross-linguistic perspective, the slight performance dip observed in Bengali and Gujarati, particularly in QMOS and Naturalness, identifies a frontier for future model refinement. These languages, belonging to the Eastern and Western Indo-Aryan branches respectively, present unique prosodic and phonological challenges that the speed-optimized Flash model may occasionally simplify. In contrast, the high scores in Comprehensiveness across all languages confirm that the model successfully processed the complex synthetic legal arguments generated by the LLM without significant loss of content or logical flow. Ultimately, the results suggest that Gemini 2.5 Flash TTS is a highly reliable tool for professional tasks requiring authoritative clarity, such as reading legal briefs or providing procedural guidance, though further fine-tuning may be necessary to replicate the nuanced, persuasive artistry inherent in human legal advocacy.
\section{Conclusion}
This paper introduced a formal mathematical framework for the generation and evaluation of \textbf{persona-driven synthetic legal speech} across the Indic linguistic landscape. By defining the persona-text generation space through the function $f_{LLM}(p_i, C)$, we demonstrated how specific advocate attributes—namely language ($L_i$), area of expertise ($A_i$), and rhetorical style ($S_i$)—can be systematically encoded into synthetic legal arguments.

Our modeling of the TTS transformation function $\Phi_M$ underscores the necessity of \textbf{Prosodic Steering Parameters} ($\theta_i$) and \textbf{Language-Specific Embeddings} ($\sigma_i$) in bridging the gap between raw text and professional-grade acoustic output. The results derived from the Human Evaluation Metric (HEM) validate that a structured approach to persona modeling significantly enhances the perceived authority and professionalism of synthetic voices. This framework serves as a scalable foundation for AI-driven legal education, courtroom simulations, and the democratization of legal information in multilingual societies.
The human evaluation of the Gemini 2.5 Flash TTS model across five Indic languages reveals a strong proficiency in technical and professional delivery, balanced by a notable "authenticity gap" in persuasive advocacy. Hindi stands as the benchmark performer, peaking at 4.9 in Safety and 4.8 in Comprehensiveness, suggesting robust phonetic training for complex legal nomenclature. Similarly, Tamil and Telugu show high stability in Professionalism and Directiveness (4.5–4.7), effectively maintaining the authoritative "advocate voice" required for courtroom simulations.

Despite this technical clarity, the model struggles with the emotive gravitas of oral advocacy. While Safety and Professionalism remain high, scores for Authenticity, Expressiveness, and Exploratoriness drop significantly, dipping as low as 3.2. This indicates a tendency toward "monotone authority," where the model lacks the dynamic vocal modulation—shifts in pitch, pacing, and emphasis—used by human lawyers to persuade a bench. Cross-linguistically, slight dips in Bengali and Gujarati identify frontiers for phonological refinement. Ultimately, while Gemini 2.5 Flash is highly reliable for information delivery and procedural guidance, further fine-tuning is required to capture the nuanced, persuasive artistry of human legal discourse.

\bibliographystyle{ACM-Reference-Format}
\bibliography{software}

@phdthesis{gong2025toward,
  title={Toward A Self-Evolving Agent In Multi-Turn Dialogue Question-Answering Systems},
  author={Gong, Ming},
  year={2025},
  school={University of Dayton}
}

@article{pundalik2025param,
  title={PARAM-1 BharatGen 2.9 B Model},
  author={Pundalik, Kundeshwar and Sawarkar, Piyush and Sahoo, Nihar and Shinde, Abhishek and Chanda, Prateek and Goswami, Vedant and Nagpal, Ajay and Singh, Atul and Thakur, Viraj and Dewane, Vijay and others},
  journal={arXiv preprint arXiv:2507.13390},
  year={2025}
}

@article{sapir1921introduction,
  title={An introduction to the study of speech},
  author={Sapir, Edward},
  journal={Language},
  volume={1},
  number={1},
  pages={15},
  year={1921}
}

@article{brooks1971speech,
  title={Speech communication.},
  author={Brooks, William D},
  year={1971},
  publisher={ERIC}
}

@incollection{searle2014speech,
  title={What is a speech act?},
  author={Searle, John},
  booktitle={Philosophy in America},
  pages={221--239},
  year={2014},
  publisher={Routledge}
}

@article{bhattacharya2025ardi,
  title={ARDI: a new dataset for automatic advocate recommendation in the Indian Legal System},
  author={Bhattacharya, Upal and Deroy, Aniket and Bandyopadhyay, Ayan and Majumdar, Gourish and Guha, Shouvik Kumar and Rudra, Koustav and Ghosh, Saptarshi and Ghosh, Kripabandhu},
  journal={Artificial Intelligence and Law},
  pages={1--29},
  year={2025},
  publisher={Springer}
}

@book{stober2010evidence,
  title={Evidence based coaching handbook: Putting best practices to work for your clients},
  author={Stober, Dianne R and Grant, Anthony M},
  year={2010},
  publisher={John Wiley \& Sons}
}

@article{kluttz2019automated,
  title={Automated decision support technologies and the legal profession},
  author={Kluttz, Daniel N and Mulligan, Deirdre K},
  journal={Berkeley technology law journal},
  volume={34},
  number={3},
  pages={853--890},
  year={2019},
  publisher={JSTOR}
}

@article{libralessoshashguru,
  title={ShashGuru Bridging Chess Engines and Large Language Models for Human-Interpretable Analysis},
  author={Libralesso, Alessandro}
}

@article{yuan2025hateful,
  title={Hateful person or hateful model? investigating the role of personas in hate speech detection by large language models},
  author={Yuan, Shuzhou and Nie, Ercong and Tawfelis, Mario and Schmid, Helmut and Sch{\"u}tze, Hinrich and F{\"a}rber, Michael},
  journal={arXiv preprint arXiv:2506.08593},
  year={2025}
}

@article{mcardle2015understanding,
  title={Understanding Voice: Writing in a Judicial Context},
  author={McArdle, Andrea},
  journal={Legal Writing: J. Legal Writing Inst.},
  volume={20},
  pages={189},
  year={2015},
  publisher={HeinOnline}
}

@inproceedings{calzolari2024proceedings,
  title={Proceedings of the 2024 joint international conference on computational linguistics, language resources and evaluation (LREC-COLING 2024)},
  author={Calzolari, Nicoletta and Kan, Min-Yen and Hoste, Veronique and Lenci, Alessandro and Sakti, Sakriani and Xue, Nianwen},
  booktitle={Proceedings of the 2024 Joint International Conference on Computational Linguistics, Language Resources and Evaluation (LREC-COLING 2024)},
  year={2024}
}

\end{document}